\documentclass{egpubl}
 
\JournalPaper         
\usepackage[T1]{fontenc}
\usepackage{dfadobe}  

\usepackage{amsmath,amsfonts}
\usepackage{algorithmic}
\usepackage{algorithm}
\usepackage{array}
\usepackage{textcomp}
\usepackage{stfloats}
\usepackage{url}
\usepackage{verbatim}
\usepackage[hidelinks]{hyperref}
\hypersetup{
  colorlinks   = true, 
  urlcolor     = blue, 
  linkcolor    = blue, 
  citecolor    = blue 
}
\biberVersion
\BibtexOrBiblatex
\usepackage[backend=biber,bibstyle=EG,citestyle=alphabetic,backref=true]{biblatex} 
\electronicVersion

\ifpdf \usepackage[pdftex]{graphicx} \pdfcompresslevel=9
\else \usepackage[dvips]{graphicx} \fi

\usepackage{egweblnk} 
\usepackage{amssymb}
\usepackage{subcaption}
\usepackage{booktabs}
\usepackage{multirow}
\usepackage[dvipsnames]{xcolor}
\usepackage{flushend}
\usepackage{marvosym}
\usepackage{cleveref}
\usepackage{xspace}
\usepackage{pdfpages}
\newcounter{rowcount}
\newcommand\sedlmair{SDR\xspace\cite{sedlmair2013empirical}\xspace}
\newcommand\datagov{Data.gov\xspace}
\newcommand\ourDataset{PSC\xspace}
\newcommand\methodName{HPSCAN\xspace}

\DeclareMathOperator{\agree}{\alpha}
\DeclareMathOperator{\pagree}{\mathcal{A}}
\DeclareMathOperator{\cross}{\kappa_{\agree}}
\DeclareMathOperator{\vanbelle}{\kappa_{v}}
\DeclareMathOperator{\simi}{\mathbb{S}}
\DeclareMathOperator{\noiseIOU}{\kappa_{n}}

\def\etal{\emph{et al}.\xspace}

\title[HPSCAN: Human Perception-Based Scattered Data Clustering]%
      {HPSCAN: Human Perception-Based Scattered Data Clustering}

\author[Hartwig \etal]
{\parbox{\textwidth}{\centering 
    S. Hartwig$^{1,*}$\orcid{0000-0001-8642-2789}
    C. v. Onzenoodt$^{1}$\orcid{0000-0002-5951-6795},
    D. Engel$^{1}$\orcid{0000-0002-5766-7215},
    P. Hermosilla$^{2}$\orcid{0000-0003-3586-4741} and
    T. Ropinski$^{1}$\orcid{0000-0002-7857-5512} 
        }
        \\
{\parbox{\textwidth}{\centering 
         $^*$corresponding author: sebastian.hartwig@uni-ulm.de \\
         $^1$Visual Computing Group, Ulm University, Germany\\
         $^2$Computer Vision Lab, Technical University of Wien, Austria
       }
}
}

\begin{document}

\teaser{
 \includegraphics[width=\linewidth]{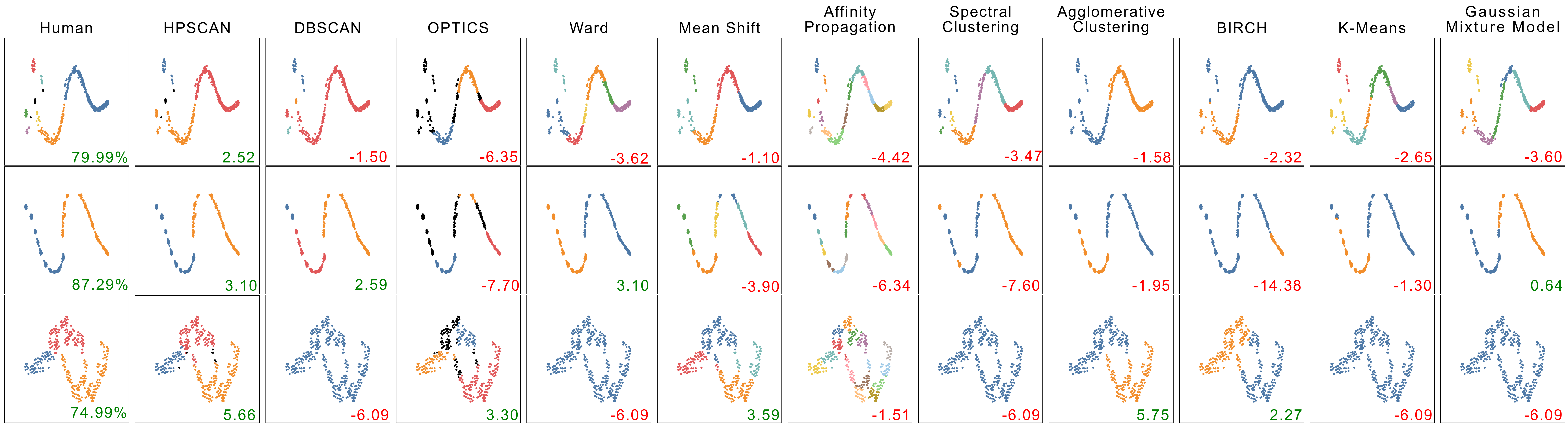}
 \centering
  \caption{\methodName clusters scatterplots in accordance with human cluster perception. Since \methodName is trained on point-based clustering data rather than images, it infers clustering  directly from the points. Here, we present a comparison of \methodName's perception-aware results against those of several state-of-the-art clustering techniques, which are optimized on our perceptual scatterplot clustering dataset (PSC). In the first column, we display the human annotation with the highest agreement rate from a group of human raters for such stimulus. We report the agreement index $\cross$, which measures whether a rating increases or decreases agreement within a group of raters. Higher values indicate better agreement.}
  \vspace*{10mm}
\label{fig:teaser}
}

\maketitle
\begin{abstract}
   Cluster separation is a task typically tackled by widely used clustering techniques, such as k-means or DBSCAN. However, these algorithms are based on non-perceptual metrics, and our experiments demonstrate that their output does not reflect human cluster perception. To bridge the gap between human cluster perception and machine-computed clusters, we propose \methodName, a learning strategy that operates directly on scattered data. To learn perceptual cluster separation on such data, we crowdsourced the labeling of $7,320$ bivariate (scatterplot) datasets to $384$ human participants. We train our \methodName model on these human-annotated data. Instead of rendering these data as scatterplot images, we used their x and y point coordinates as input to a modified PointNet++ architecture, enabling direct inference on point clouds. In this work, we provide details on how we collected our dataset, report statistics of the resulting annotations, and investigate the perceptual agreement of cluster separation for real-world data. We also report the training and evaluation protocol for \methodName and introduce a novel metric, that measures the accuracy between a clustering technique and a group of human annotators. We explore predicting point-wise human agreement to detect ambiguities. Finally, we compare our approach to ten established clustering techniques and demonstrate that \methodName is capable of generalizing to unseen and out-of-scope data.


\end{abstract}  

\section{Introduction}
Clustering is often applied to bivariate projections of multidimensional data visualized as scatterplots, which aids humans in identifying cluster patterns within a large set of data points~\cite{cavallo2019clustrophile2}. The insights gained from this process can guide further analysis or decision-making. The effectiveness of clustering depends on various factors, such as the choice of the clustering algorithm, its parameters, and the number of clusters to be identified, the development of cluster algorithms has been researched for decades~\cite{ester1996density,zhang1997birch,ankerst1999optics}. Although these algorithms achieve meaningful separation of clusters, the resulting clusters often do not correlate with the perceptual cluster separation of the human visual system, as illustrated in \Cref{fig:teaser}. They also require careful hand-tuning of parameters~\cite{aupetit2019toward} or reliance on external validation measures~\cite{liu2010understanding}. Recent works even regard human visual perception as the gold standard for evaluating clustering~\cite{xia2021visual}.

The discrepancy between clustering algorithms and the perceptual cluster separation of the human visual system can hinder humans' ability to interpret clustering results~\cite{aupetit2019toward}. When algorithms identify clusters that do not align with the way humans would naturally group data, it becomes more challenging to understand and interpret the patterns and relationships within the data~\cite{etemadpour2014role} or trust the computational clustering model~\cite{sacha2015role}. Aligning clustering outcomes with human perception significantly enhances the interpretability and the applicability of results, ensuring that decisions are based on insights that are both computationally valid and intuitively understandable. In some instances, intuition may outperform rational or analytical reasoning~\cite{sadler2004intuitive, dane2012should}. This alignment is especially beneficial in applications where decision-makers heavily rely on visual data interpretation to make informed choices~\cite{huang2018role,luan2019ecological}. Moreover, existing clustering algorithms lack the ability to indicate the confidence of their output. Therefore, we propose \methodName, a learning-based clustering model, that through training on empirical data, integrates the way humans naturally group and detect data. It can automatically quantify grouping patterns in 2D scattered data, separate outliers from clusters, and provide human agreement scores for such estimations.

Our goal is for \methodName to be applied directly to point-based data, rather than to scatterplot images. This enables researchers to comfortably interchange existing clustering techniques with \methodName. Unlike image-based approaches, such as ScatterNet~\cite{ma2018scatternet}, our approach focuses solely on the human-aligned clustering of point-based data and does not require the generation of scatterplot images. Learning to predict point-wise cluster assignments from human-annotated point data renders standard classification losses inapplicable due to the arbitrary orderings of cluster IDs in the annotations. Therefore, we adapt meta-classification learning for segmentation tasks, a technique proposed by Hsu \etal\cite{hsu2015kcl, hsu2019mcl}, \methodName exhibits this property. Similarly, the order of point data can be arbitrary, requiring the ability of input order invariance. By adapting the PointNet++~\cite{qi2017pointnet++} as base architecture \methodName exhibits such property. To evaluate our model's predictions, we need to measure its agreement with human raters. The standard Vanbelle Kappa Index~\etal\cite{vanbelle2009agreement}, used to measure agreement between a single rater and a group of raters, does not work in certain atypical cases that are still frequent in our datasets. This has led us to develop an alternative metric. The collected dataset not only enables us to investigate human rater agreement but also to incorporate this knowledge into the model predictions.
Within this paper, we make the following contributions:

\begin{itemize}
    \item We propose \methodName, a point-based neural clustering algorithm that clusters point-based data following human cluster perception.
    \item We present a novel loss function that accounts for point-to-cluster matching invariant to cluster IDs.
    \item We release a large-scale point-based clustering dataset, containing $7,320$ human annotated scatterplots from real-world data\footnote{The data that support the findings of this study are openly available at \url{https://github.com/kopetri/HPSCAN}}.
    \item We predict human uncertainty in different usage scenarios.
    \item We introduce an outlier-aware, generalizable metric for measuring agreement between an isolated rater and a group of raters.
    \item We explore the quality of \methodName compared to competitors, sensitivity to various aspects, and generalization to new datasets.
\end{itemize}

\section{Related Work}\label{sec:related_work}
Many approaches for clustering bi-variate data have been developed in the last few decades. In this section, we provide an overview of existing approaches, first focusing on algorithms based on hand-crafted features and then discussing learning-based algorithms. Then, we will discuss approaches which take into account human perception. Finally, we provide an overview of existing scatterplot datasets for which human judgments were collected.

\noindent\textbf{Conventional clustering.} Over the years, a vast array of algorithms for clustering data has been developed. These include density-based clustering~\cite{ester1996density}, hierarchical clustering~\cite{moseley2017approximation}, subspace clustering~\cite{elhamifar2013sparse}, fuzzy clustering~\cite{bezdek1984fcm}, and co-clustering~\cite{dhillon2003information}, with new approaches emerging annually. While some methods have a predetermined maximum number of clusters, others require tuning to determine appropriate values for neighborhood size, point distance threshold, or bandwidth. One of the most widely used methods, DBSCAN~\cite{ester1996density}, assesses distances between the nearest points and identifies outliers. Zhang \etal\cite{zhang1997birch} propose BIRCH, a tree-based cluster algorithm, that uses an existing agglomerative hierarchical clustering algorithm to cluster leaf nodes. Ankerst \etal\cite{ankerst1999optics} presented a density-based clustering algorithm that generates a cluster order, reflecting the cluster structure of a given set of points.  

\noindent\textbf{Optimization-based Clustering.} One of the earliest clustering techniques is K-Means\xspace\cite{hartigan1979algorithm}, which aims to find a partition that minimizes global within-cluster variance. This was followed by the introduction of learnable clustering algorithms, such as Neural Gas \cite{martinetz1993neural}. Du \etal\cite{du2010clustering} and Schnellback \etal\cite{schnellbach2020clustering} provide comprehensive overviews of neural clustering approaches. Self-organized maps (k-SOM), developed by T. Kohonen~\cite{kohonen1991som,ghaseminezhad2011novel,amarasiri2004hdgsom}, represents a neural clustering algorithm that operates on a grid of neurons. In this algorithm, the network learns to assign clusters to proximal data points that are close to each other. There are several surveys on clustering using deep learning approaches\xspace\cite{Aljalbout2018ClusteringWD,karim2020deep,Ren2022DeepCA,zhou2022comprehensive}, which describe techniques for utilizing feature encoding to learn representations conductive to clustering. However, these works primarily focus on optimizing clusters in an unsupervised manner, where clusters are defined by computed quantities and are enforced through cluster loss functions, rather than by human judgments. This work focuses on the development of a deep learning method that is supervised by humans, with the goal of identifying clusters in the underlying data.

\noindent\textbf{Interactive Clustering.} Numerous approaches have been developed for interactive cluster analysis\xspace\cite{imre2007clustersculptor,lex2010matchmaker,metsalu2015clustvis,BRUNEAU2015627,Demiralp2017ClustrophileAT}, offering users control over the dimension reduction techniques, visual encoding, cluster parameters, and more. Xia \etal\cite{xia2022interactive} propose an interactive cluster analysis method using contrastive dimensionality reduction. Initially, a neural network generates an embedding to reduce the dimensionality of a given high-dimensional dataset. Subsequently, users interactively select data points to establish 'must-link' and 'cannot-link' connections between clusters. The neural network is then re-trained using contrastive learning to update the embedding. Additionally, Cavallo et al.\cite{cavallo2019clustrophile2} introduce \textit{Clustering Tour}, an interactive tool designed to assist users in selecting clustering parameters and evaluating the effectiveness of various clustering outcomes in relation to their analysis goals and expectations.

\noindent\textbf{Image-based Machine Learning Methods for Clustering}
Image-based deep learning has been successfully applied to bivariate data by learning from scatterplot images. In the work of Pham \etal\cite{pham2020scagcnn}, a CNN is trained to estimate visual characterizations of scatterplots. Xia \etal\cite{xia2021visual} propose a CNN-based approach to analyze the factors of class and cluster perception based on human judgement data. Fan \etal\cite{fan2018fast} proposes a CNN trained on density histogram images and human-generated brushing of underlying scatterplots to automatically brush areas in scatterplots, without selecting individual points, which is a selection-targeted clustering technique. These image-based approaches rely on the projection of bivariate data onto a 2D canvas. While in many scenarios scatterplots might be readily available or straightforward to generate, problems such as overdraw can arise when visual encoding is not selected properly. \methodName infers clusters from point data directly, without the need to generate images.

\noindent\textbf{Point-based Machine Learning Methods for Clustering} In the past years, several point-based learning architectures have been proposed that directly learn from point cloud data~\cite{qi2017pointnet, qi2017pointnet++,hermosilla2018monte, jiang2020end, ding2019votenet}. Chen\etal\cite{Chen_2019_CVPR} proposes a deep hierarchical cluster network called ClusterNet, which better adapts to rotations of 3D objects. It is argued that existing data augmentation strategies or rotation equivariant networks cannot guarantee to satisfaction of all rotation-equivariant constraints at each layer. Therefore, ClusterNet introduces a rigorous rotation-invariant representation by employing hierarchical clustering to explore and exploit the geometric structure of the point cloud. In the work of Wöhler \etal\cite{wohler2019learning}, PointNet++ was demonstrated to be effective for 2D point cloud learning, estimating the correlation of data dimensions visualized by scatterplots from human-annotated data. 

\noindent\textbf{Perception-based clustering.} Most of perception-based approaches have been proposed for visual quality measures. Quadri \etal\cite{quadri2020modeling} crowdsource cluster counts from human observers for synthetic scatterplots. They use distance and density-based algorithms to compute cluster merge trees. Furthermore, they utilize the merge trees with a linear regression model to predict the number of clusters a human would perceive in a scatterplot, without identifying the actual point-to-cluster assignment. Abbas \etal\cite{abbas2019clustme} propose ClustMe, a visual quality measure to rank scatterplots based on the complexity of their visual patterns. It encodes scatterplots using Gaussian Mixture Models (GMM) before optimizing a model of component-merging based on human judgments. ClustML~\cite{ullah2021clustml} enhances ClustMe by utilizing an automatic classifier trained on human perceptual judgments, replacing heuristic-based merging decisions with more accurate, data-driven models that align more closely with human visual assessment. 
While these approaches are GMM-based quality measures ranking scatterplots based on their grouping patterns, we propose a clustering algorithm that assigns cluster labels to individual points.
A recent approach called CLAMS, introduced by Jeon \etal\cite{jeon2024clams}, measures cluster ambiguity in visual clustering. They trained a regression model using handcrafted features based on perceptual data to estimate a score representing cluster ambiguity of an input scatterplot. In contrast, \methodName learns to represent features of a scatterplot implicitly based on its point encoder.
Furthermore, Sedlmair and Aupetit~\cite{sedlmair2013empirical,sedlmair2015data} evaluated $15$ visual quality measures for class separability of labeled data based on human judgments. These measures only apply to already labeled data points as a whole and do not allow for the alignment of individual points to clusters. ScatterNet, proposed by Ma \etal\cite{ma2018scatternet}, is a learned similarity measure that captures perceptual similarities between scatterplots to reflect human judgments. Aupetit \etal\cite{aupetit2019toward} provide an evaluation of $6$ state-of-the-art clustering techniques on a perception-based benchmark~\cite{abbas2019clustme}. In their evaluation, they assess a cluster counting task, where the benchmark provides human decisions for a scatterplot on whether one or more than one cluster were perceived. They can show that agglomerative clustering techniques are in substantial agreement with human raters. However, in this work, we asked human raters for point-wise cluster decisions, rather than a binary decision on the number of clusters.

\begin{table}[t]
\setlength{\tabcolsep}{1.5pt}
\centering
\footnotesize
\caption{Overview of existing scatterplot datasets featuring subjective human judgments. Judgments collected for our dataset consists of richly annotated scatterplots, rather than binary decisions.}\label{tab:human-judgement-datasets}
\resizebox{\columnwidth}{!}{%
\begin{tabular}{lcccll}
\toprule
Dataset & participants & responses & stimuli & modality & human judgment \\
\midrule
ScatterNet~\cite{ma2018scatternet} & 22 & 5,135     & 50,677 & real      & similarity perception \\
ClustMe~\cite{abbas2019clustme}    & 34 & 34,000    & 1,000  & synthetic & cluster count (binary) \\
ASD~\cite{quadri2022automatic}     & 70 & 1,259     & 5,376  & real      & visual encoding\\
VDCP~\cite{quadri2020modeling}     & 26 & 1,139     & 7,500  & synthetic & cluster count \\
HSP~\cite{pandey2016towards}       & 18 & 4,446     & 247   & real      & similarity perception \\
WINES~\cite{tatu2010visual}        & 18 & 90       & 18    & real      & class separability \\
SDR~\cite{sedlmair2013empirical}   & 2  & 1,632     & 816   & real+synth& class separability \\
VCF~\cite{xia2021visual}           & 5  &  152K     & 50,864  & real+synth & binary cluster separation \\ 
CLAMS~\cite{jeon2024clams}         & 18 & 1,080      & 60    & synthetic & point-wise cluster separation \\
\bottomrule                                             
\ourDataset                         & 384 & 7,320    & 1,464   & real      & point-wise cluster separation
\end{tabular}%
}
\end{table}

\noindent\textbf{Scatterplot datasets.} Our provided dataset is not the first scatterplot dataset available to the research community. Existing work has already investigated subjective human judgments in the context of scatterplots~\cite{ma2018scatternet,abbas2019clustme, quadri2022automatic, quadri2020modeling, quadri2021survey, pandey2016towards, tatu2010visual, sedlmair2013empirical}. As listed in \Cref{tab:human-judgement-datasets}, datasets have been collected featuring diverse human judgments, such as similarity perception, class separability, or cluster counts. Sources for real scatterplots include popular datasets like MNIST, Rdatasets~\cite{rdatasets}, or scatterplots synthetically generated based on Gaussian Mixture Models. While these datasets provide valuable human judgments for scatterplots, they lack complexity, as point-wise judgments are crucial for investigating human cluster perception. Therefore, we have collected a point-annotated cluster dataset, for which we describe the crowdsourcing process and provide statistics in the following subsections.

\noindent\textbf{Prerequisite.} \methodName is based on PointNet++, which we detail further for the reader in this paragraph. PointNet++ is a hierarchical neural network that effectively captures local geometric structures in point data. It introduces set abstraction layers composed of three fundamental operations: sampling, grouping, and feature extraction. In the sampling step, Farthest Point Sampling (FPS) is employed to select a subset of representative points from the input, preserving crucial geometrical features while reducing computational complexity. The grouping step organizes neighboring points into local regions centered around each sampled point. For feature extraction within these local neighborhoods, PointNet++ utilizes 1D convolutions to compute point-wise features. These features are then aggregated using symmetric functions such as max pooling, ensuring that the network remains invariant to the order of points, as the aggregation does not depend on the sequence of points. This hierarchical and permutation-invariant design enables PointNet++ to effectively learn both local and global features, making it a suitable point encoder for tasks such as classification and segmentation in point data.

\section{\methodName}\label{sec:method}
To learn and imitate human cluster perception based on point-based clustered data, we propose \methodName, a model that clusters point-based data using human annotations. In this section, we will first discuss the design choices for our approach (see Subsection~\ref{ssec:design-considerations}). Then, we will describe the architecture of \methodName (see Subsection~\ref{subsec:architecture}), before discussing the loss function used for training \methodName (see Subsection~\ref{subsec:loss}).

\subsection{Design Considerations}\label{ssec:design-considerations}
Learning to cluster bivariate data is a classification task where a deep learning model takes as input a vector of N 2D points and produces as output another vector of real numbers that assign each point to one of K classes. We must ensure invariant to input permutation, as different point orderings yield the same scatterplot, and invariance to class assignment, as the identity of the class does not matter. \\
\newline
\textbf{Clustering Technique.} \methodName is developed as a clustering technique with the goal of mirroring human perceived clustering on monochrome scatterplots. Therefore, the input to our model is a set of 2D points. \methodName assigns each point a cluster ID indicating cluster affiliations. Human judgments for clustering scatterplots are not identical between individual subjects for the same scatterplot, as shown in Section~\ref{ssec:exp_human_agreement_maximization}. Consequently, \methodName estimates for each point a human agreement score indicating the degree of agreement between human raters. In detail, our approach estimates $N$ pairs $(C, A)$ with cluster ID $C$ and agreement score $A$ for a given set of $N$ points. \\
\newline
\textbf{Point-wise clustering.} While clustering by human raters is performed in the image domain through visualization of scattered data, one could argue for developping an image-based clustering technique to imitate human-perceived clustering. We assume that data are already available as pointsets before being encoded as scatterplot. Thus, we design \methodName to operate on a point-level basis assigning cluster IDs to each point directly. \\ 
\newline
\textbf{Applicable to unstructured data.} \methodName must be able to learn from unstructured data. Unfortunately, learning from such unstructured data poses several challenges. First, the ordering of our scattered point data, as stored on disk, might vary without actually affecting the visualization. Such effects are sometimes overlooked in the visualization community but have recently also been investigated for line graphs by other researchers~\cite{trautner2021line}. For training \methodName, we require invariance over point ordering. Another challenge is that our visual stimuli are often sparse, meaning they contain a large number of empty regions, which makes more challenging to extract useful information from the data.\\
\newline
\textbf{Uncertainty.} Since crowdsourcing annotations involve human raters, our collected annotations per scatterplot may contain arbitrary cluster assignments, especially in cases where no clear patterns or shapes are present. Thus, we design \methodName to output an agreement score that indicates uncertainty. \\
\newline
\textbf{Order-invariance.} When predicting clusters, we want \methodName to be invariant to the cluster IDs, since cluster IDs can be permuted without affecting the correctness of the result. Therefore, we need to implement a model architecture and a loss function that are invariant to these orderings. We utilize meta-classification learning, which is invariant to cluster ID order.

\subsection{Model Architecture}\label{subsec:architecture}
\begin{figure*}
    \centering
    \includegraphics[width=\textwidth]{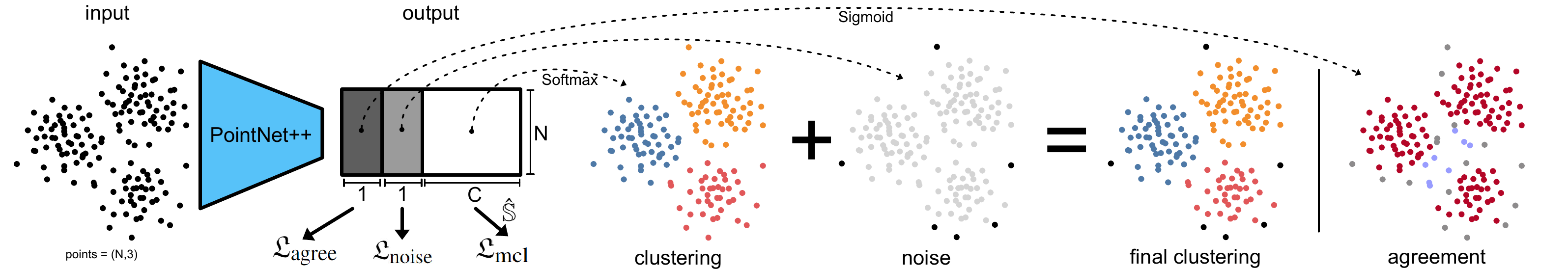}
    \caption{Overview of the input and output parameters of \methodName: A point set of $N$ points is processed by a PointNet++, which outputs for each point three outputs: agreement score, noise and cluster affiliations. The latter two outputs are combined to produce the final point-wise clustering estimation.}
    \label{fig:architecture}
\end{figure*}


To address the challenges outlined above, we decided to adapt the PointNet++ architecture~\cite{qi2017pointnet++} to learn from our point-based scatterplot dataset. PointNet++ has been demonstrated to perform well in classification and segmentation tasks, and it has been applied to 2D point clouds sampled from MNIST~\cite{lecun1998mnist} images. Following their example, we fix the input size to $N=512$ points in Euclidean space and set the z-axis of all points to $0$ which transforms the model into a 2D architecture as all values of computations and weight matrices along the third dimensions are removed. The feature extractor consists of $4$ hierarchical layers, for both down-sampling and up-sampling stages, where we use point cloud sizes $256, 128, 64, 8$ for farthest point sampling (FPS). Note that we use a randomly initialized FPS during training and fix it for inference. Additionally, we change the implementation of PointNet++, that for each point, the four feature propagation stages of PointNet++ have output feature sizes $256, 256, 128, 128$. Finally, the point encoder projects feature vectors of size $128$ using a final 1D convolution layer, outputting an agreement score $\agree$, cluster probability $P_c$, and noise probability $P_\text{noise}$ for each point. The output of our network has the shape $Nx(C+2)$, where $C$ is the maximum number of clusters, see \Cref{fig:architecture}.

\subsection{Training Loss}\label{subsec:loss}
\begin{figure}
    \centering
    \includegraphics[width=\linewidth]{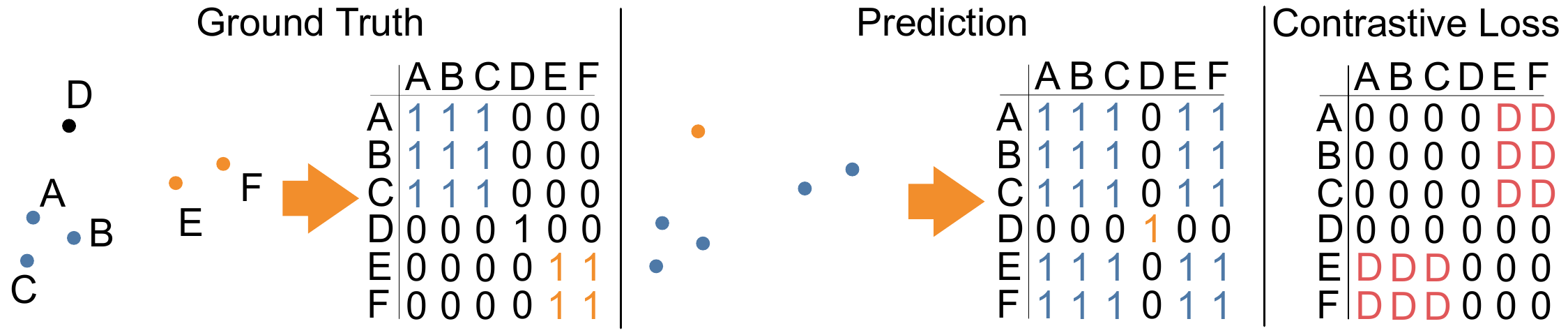}
    \caption{Visualization of the encoding of ground truth and predicted clusters, along with the corresponding contrastive loss. Scattered data with color coded clustering (\emph{left}), and corresponding similarity matrix $\simi$, see \Cref{eq:similarity_matrix} (\emph{right}).}
    \label{fig:similarity_matrix}
\end{figure}

Commonly used loss functions, such as negative log-likelihood, would require a fixed cluster order and are therefore not applicable. This is because they would imply that the position of a cluster is related to the number of clusters present in the target. Specifically, we need a loss function that computes identical gradients for predictions $A=[c_0, c_1, c_2]$ and $B=[c_2, c_0, c_1]$ given that there are three clusters in the ground truth. Without this, the model would chase arbitrary gradients, resulting in a lack of convergence, which underscores the importance of a specialized loss function. Therefore, we draw from the field of meta-classification learning to use a contrastive loss, as proposed by Hsu \etal\cite{hsu2015kcl,hsu2017learning,hsu2019mcl}. Meta-classification learning addresses a multi-class problem by reformulating it as a binary-class problem. It optimizes a binary classifier for pairwise similarity prediction and through this process learns a multi-class classifier as a submodule. Consequently, we represent our cluster targets as a similarity matrix $\simi$, which encodes point-wise cluster affiliation. It has the form $\simi \in N \times N$, where $N$ is the number of points in the scatterplot, and it is defined as:
 
\begin{equation}
    \simi_{i,j}(C) = \begin{cases}
         1, & C_i, C_j~\text{in the same cluster (positive)} \\
         0, & \text{in different clusters (negative)}
    \end{cases}
    \label{eq:similarity_matrix}
\end{equation}

\noindent where $C$ represents the point-wise cluster IDs for $N$ points. This matrix contains positive values for similar points that belong to the same cluster and zero values for points that are dissimilar, i.e., not in the same cluster. For clarification, we demonstrate the representation of our targets in \Cref{fig:similarity_matrix}. It is evident that permuting the order of points results in the same matrix with equal permutation applied.

We define the loss term based on this adjacency matrix representation, which does not require the identification of clusters, thereby making it invariant to cluster identity, as follows:

\begin{equation}
    \mathfrak{L}_\text{mcl} = -\sum_{ij}{\omega_{D}^{ij}(\simi_{ij}\, \log\,\hat{\simi}_{ij}\,+\,(1-\simi_{ij})\,\log(1-\hat{\simi}_{ij})})
    \label{eq:mcl_loss}
\end{equation}

\noindent where $\omega_{D}$ is a weight matrix used to rescale the influence of negative samples in our contrastive loss, and $\hat{\simi}$ is the estimated similarity matrix, which can be derived from the output. Furthermore, we use it to correct the imbalanced error contribution from small clusters by applying an increasing weight factor to the loss function. We define this weight matrix as follows:

\begin{equation}
    \omega_{D}^{ij}(\simi) = \begin{cases}
        \frac{1}{w_c^2}, & \simi_{ij} = 1 \\
        D,               & \simi_{ij} = 0 \\
    \end{cases}
\end{equation}

\noindent where $D$ is the momentum of negative samples and $w_c$ is the cluster-specific contribution, defined as $w_c=N_c/N$. Here, $N_c$ is the number of points for a specific cluster $c$, and $N$ is the total number of points. 

Separating clusters in scattered data is not always well-defined. While it might be obvious in some cases, in others, points may not cluster at all, or outliers might be present in the scatterplot. In the following, we treat non-clustering points and outliers the same, referring to both as \textsc{noise}. To better capture the aspect of \textsc{noise}, we introduce another loss term $\mathfrak{L}_\text{noise}$. It is computed using a weighted binary cross entropy loss to counter the strong class imbalances between positive (\textsc{cluster}) and negative (\textsc{noise}) cases. In our experiments, we found that scaling the weighting of negative cases by $9$ is beneficial. We reserve cluster ID $0$ for points annotated as \textsc{noise}. To convert a multi-cluster target, which contains multiple cluster IDs, into a binary target (0 = noise and 1 = not noise), we replace all cluster IDs greater than $0$ with $1$. The loss term is then defined as follows:

\begin{equation}
    \mathfrak{L}_\text{noise} = -\frac{1}{N}\sum_{i}^{N}{0.9\,\nu_i\,\log(\hat{\nu}_i)+0.1\,(1-\nu_i)\,\log(1-\hat{\nu}_i)}
    \label{eq:noise_loss}
\end{equation}

where $\nu_i$ is the binary ground truth label, and $\hat{\nu}_i$ is the predicted probability of point $i$ being classified as noise.

Finally, we aim to estimate an agreement score using \methodName. To compute the agreement between two clusterings, $R_a$ and $R_b$, obtained from raters a and b for the same point set, we propose the following measure, referring to it as the agreement score $\pagree$. We define it for a single point $i$ as follows:

\begin{equation}
    \pagree_i(R_a, R_b) = \frac{1}{N}\sum_{j}^{N}{1 - |\simi_{ij}(R_a) - \simi_{ij}(R_b)|}
    \label{eq:agreement_score}
\end{equation}

\noindent where $R_a$ and $R_b$ are the annotations of two participants for the same scatterplot $R$. The group $G$ of raters from our study is a sample of the population of all possible human raters we aim to model. We can estimate what would be the agreement among any pairs of raters in the general population by averaging the $K = \binom{M}{2}$ pairwise rater agreements taken among the $M$ raters we have in $G$:

\begin{equation}
    \agree_i(G) = \frac{1}{K}\sum_{k}^K{\pagree_i(R_{a_k}, R_{b_k})}\\
    \label{eq:average_agreement_score}
\end{equation}

Finally, training \methodName to predict human agreement is done by adding the agreement loss term to \Cref{eq:total_loss}, which is defined by:

\begin{equation}
    \mathfrak{L}_\text{agree} = \frac{1}{N}\sum_i^N{| \agree_i - \hat{\agree}_i |}
    \label{eq:agreement_loss}
\end{equation}
where $\agree_i$ is the aggregated agreement score between all human raters, see \Cref{eq:average_agreement_score} and $\hat{\agree}_i$ is the prediction by \methodName.

Now, we can combine all loss terms using a sum, whereby our loss term $\mathfrak{L}_\text{total}$ is composed by rescaling $\mathfrak{L}_\text{mcl}$ by a factor of $0.1$ adjusting it to similar loss contribution. This results in the following overall loss:

\begin{equation}
    \mathfrak{L}_\text{total} = 0.1\mathfrak{L}_\text{mcl}\,+\mathfrak{L}_\text{noise}+\mathfrak{L}_\text{agree}
    \label{eq:total_loss}
\end{equation}

\subsection{Training Protocol}\label{ssec:train_protocol}
To train \methodName, we use the Adam optimizer~\cite{kingma2014adam}, with $\text{betas}=(0.9,0.999)$ and a learning rate of 1e-5, reducing the learning rate by a factor of $10$ when the validation loss stagnates for $50$ epochs, with a batch size of $32$. During training, we apply random data augmentation to the input point clouds, including horizontal and vertical flips, and random rotations between $-180$ and $180$ degrees. We then normalize all points clouds to be in the range of $-1.0$ to $1.0$, centered around the origin, identical to the procedure described in \Cref{ssec:stimuli_collection}. Additionally, we adopt a random crop transformation, which performs random horizontal and vertical cuts through the points. Points from the left (top) side of the cutting line are moved to the right (bottom) side, and vice versa. We keep track of the clusters that get cut, thereby increasing the number of cluster annotations accordingly. For details on how we implement this data augmentation strategy, we refer the reader to Section 6 in the supplemental material.

\section{Perceptual Scatterplot Clustering Dataset}\label{sec:dataset}
This section provides details on the collection of our perceptual scatterplot clustering dataset (\ourDataset), which is used to train \methodName. To collect a large number of annotations suitable for training \methodName, we crowdsourced annotations for our dataset online using \textit{Prolific}.

\subsection{Stimuli Selection}\label{ssec:stimuli_collection}
The correct selection of stimuli is crucial for collecting high-quality annotations. It was essential to choose real-world stimuli, rather than more simplistic ones generated with Gaussian mixture models, to test our approach in real-world scenarios. Therefore, we downloaded data from \url{https://data.gov}, the United States government's open data website, which offers access to datasets covering a broad range of topics, including agriculture, climate, crime, education, finance, health, energy, and more. At the time of collecting the dataset, the site provided access to more than $240,000$ datasets. We chose to collect only data available as CSV files for easier processing. In the first preprocessing step, we filtered out datasets with fewer than $512$ or more than $10,000$ rows to generate comparable stimuli. Additionally, we used only CSV files with more than two columns, as we require at least two data dimensions to create two-dimensional plots. For datasets with more than $512$ rows, we randomly sampled a fixed number of $512$ rows. Following the approach of Sedlmair \etal\cite{sedlmair2013empirical}, we applied dimensionality reduction techniques to these datasets, yielding $1,464$ scatterplots. For dimensionality reduction, we used t-SNE~\cite{van2008visualizing} and PCA~\cite{jolliffe2002principal} from the scikit-learn framework~\cite{pedregosa2011scikit}, using default parameters. Finally, we normalized all datasets by centering the points around $(0,0)$ and applying minimum/maximum normalization, with positions lying within the range of $[-1, 1]$. The resulting dataset has been used to generate stimuli for our crowdsourcing process.

\subsection{Crowdsourcing Process}\label{ssec:online_study}
In the past, crowdsourcing experiments have proven useful for collecting large amounts of annotated data~\cite{hartwig2022learning, heer2010crowdsourcing, buhrmester2011amazon, van2022out, yang2023how}. To crowdsource our training data from a large group of raters, we built a web-based framework supporting mouse and keyboard interactions. Crowd workers were exposed to this framework and tasked with segmenting clusters in scatterplots generated from the stimuli selected as discussed above. For intuitive interaction, clusters had to be segmented by brushing points, with the brush color varied per cluster.

The entire crowdsourcing process is divided into three parts. First, each crowd worker received an introduction, where they watched a 3-minute video describing the task, example stimuli with corresponding clustering, and examples of good and bad clustering. We refer the reader to the supplementary material for the tutorial video. To segment the scatterplots on a point-based level, we presented the plots vector-based, rendered on a $500\times 500$ pixel-sized canvas using a marker size of $5$ pixels. We allowed participants to brush the points using a circular brush with variable size, enabling them to make both coarse and fine-grained annotations. Initially, all points within the scatterplot are colored black; during the segmentation process, participants use the brush to colorize the points within a cluster using a selected color. Initially, the interface provides only a single color for brushing points. However, participants can add more colors (up to $20$) by clicking a $+$ button. If participants were unable to delineate any clusters, we enabled them to explicitly state this using a checkbox labeled \textit{I can't see any cluster}. Participants were further instructed that points without cluster affiliation should remain black, signaling that they are considered outliers or noise. Once confident with a segmentation, participants pressed a button below the plot to continue to the next stimulus.

In the second part of the crowdsourcing process, each participant practiced the task during a short training phase and received performance feedback. While this part used stimuli not included in the main study, these stimuli formed obvious clusters, chosen to align with the expected agreement between clustering algorithm results and capable human observers. Accordingly, we used this clustering as ground truth for the participant feedback. We could also verify participant performance through these stimuli, allowing participants to continue to the study's final part only if they completed all training stimuli successfully.

In the third part of the crowdsourcing process, participants annotated clusters in $20$ stimuli each during the main study. To detect bots or \textit{click-through} behavior, we added three additional sanity checks. Such a stimulus displays multiple spatially separated Gaussian blobs forming visually distinct clusters. We present such a stimulus used as a sanity check, for which we predefined ground truth cluster separation. Participant segmentations that diverged more than $30\%$ from the target failed this sanity check, and we discarded the data from participants who failed more than one sanity check. In Section 1 of the supplemental material, we show the user interface of the web application used, along with an example of a sanity check.

Using this procedure, we collected $7,320$ clusterings for $1,464$ scatterplots from $384$ participants, with each stimulus annotated by at least $5$ individuals. On average, a participant took $15.1$ minutes to complete our study, with $257$ male, $125$ female, and $2$ participants who preferred not to provide their sex. The average age of the participants was $32.7$ years. We had to reject a single participant for failing the sanity checks and an additional $17$ participants for discontinuing the study. In the following, we report statistics regarding the collected dataset. Compared to existing datasets, see \Cref{tab:human-judgement-datasets}, our study obtained annotations from $5$ times more subjects than previous dataset studies.

\subsection{Annotation Analysis}\label{ssec:human_cluster_results}
\begin{figure}[t]
    \centering
    \includegraphics[width=\linewidth]{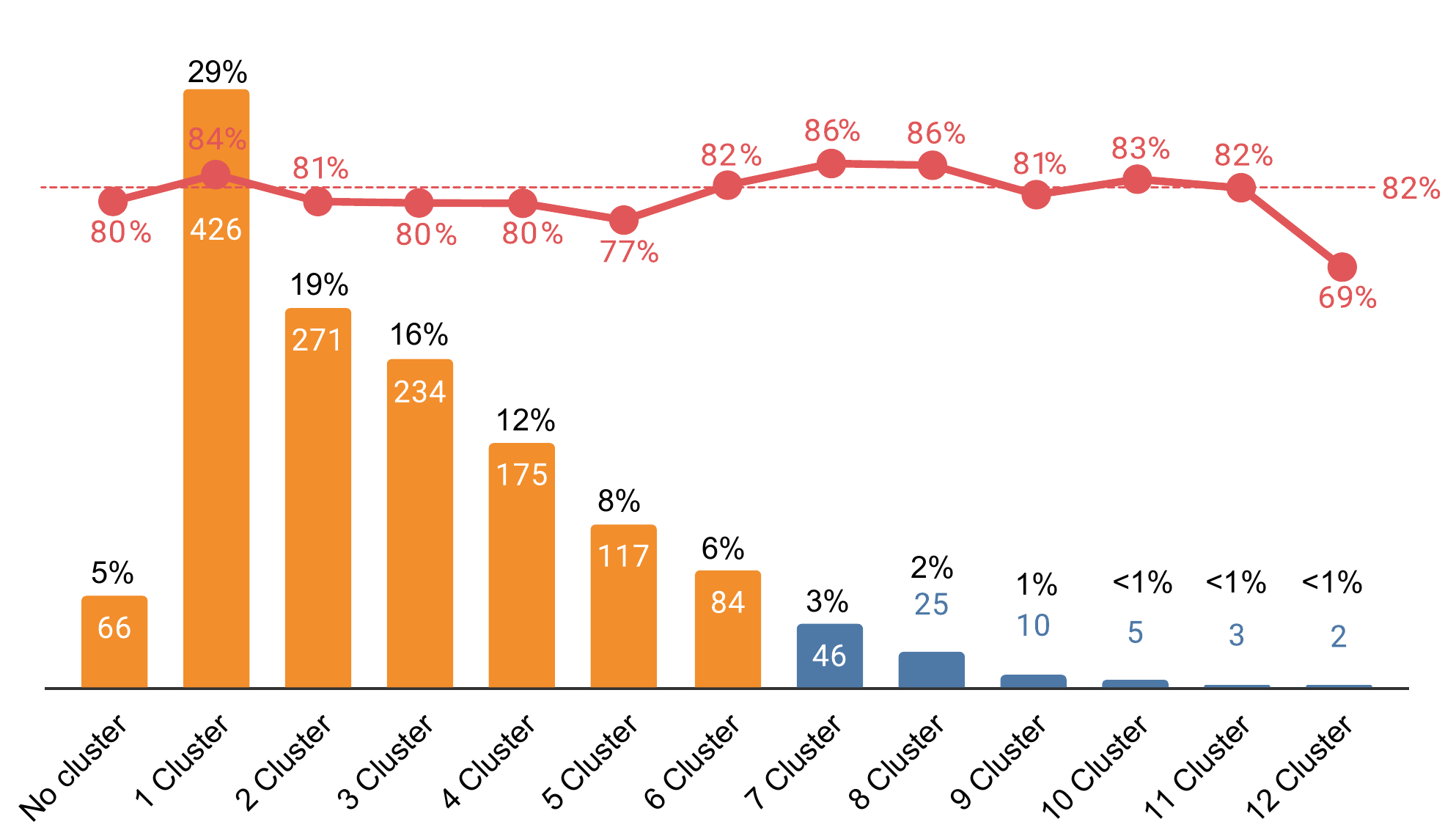}
    \caption{Our dataset, \ourDataset, consists of $1,464$ stimuli. We show the distribution of stimuli that were annotated with a certain number of clusters. Rater agreement is visualized in red for each number of clusters. The white numbers correspond to the number of stimuli for each cluster count, while the black numbers indicate the corresponding amounts. The red line represents the average agreement for the entire dataset.}
    \label{fig:dataset_statistic}
\end{figure}

To investigate the agreement between human judgments, we computed an agreement score $\agree$, which is defined in~\Cref{sec:metrics}. The average agreement score for \ourDataset is $81.9\%$. In~\Cref{fig:dataset_statistic}, we display the average agreement scores stratified by the number of clusters. The number of clusters is determined by checking if more than half of the participants agree on the same number of clusters. If this agreement cannot be found, we compute the average number of clusters from all five annotations. The results show that the participants more often annotated fewer clusters, as indicated by the decreasing number of annotations with an increasing number of clusters.

\begin{figure}[b]
    \centering
    \includegraphics[width=\linewidth]{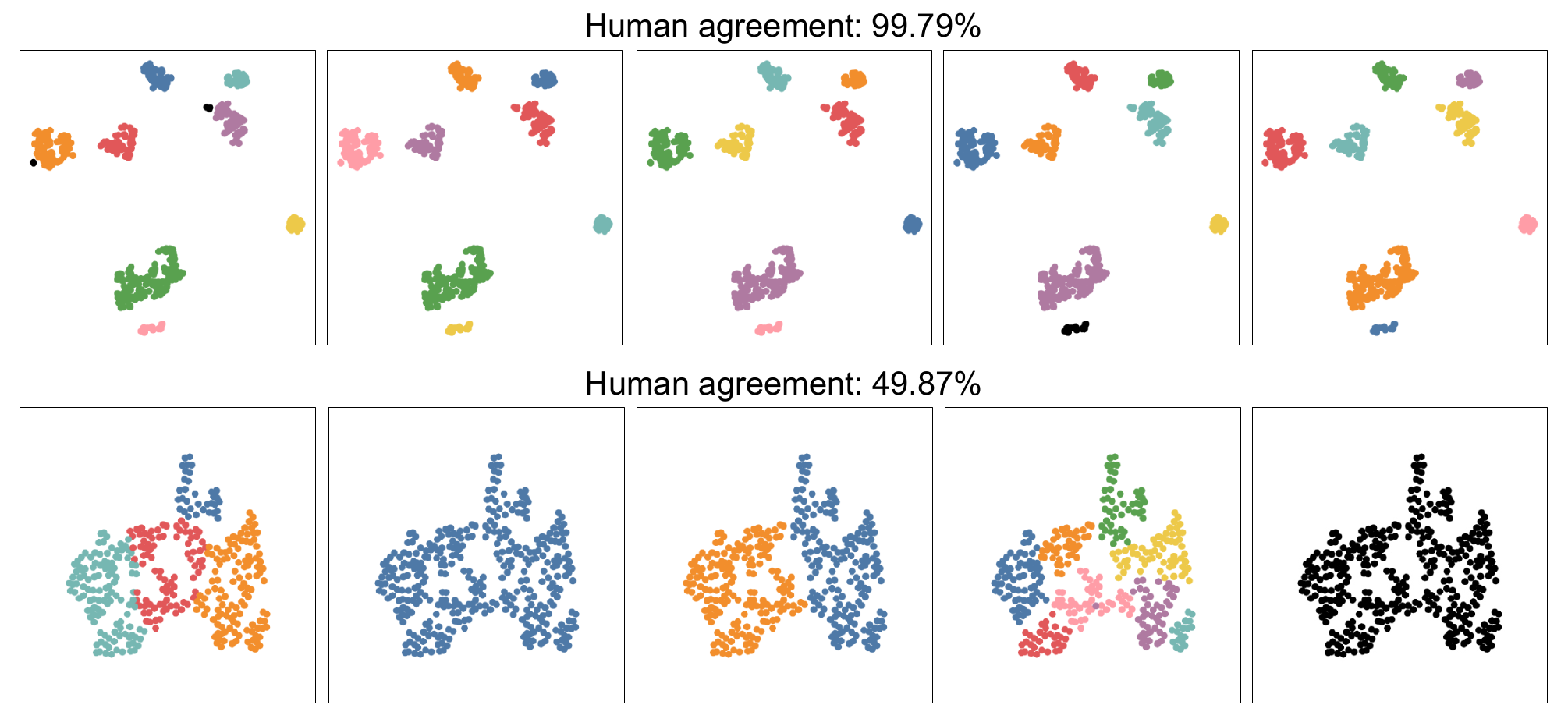}
    \caption{Comparison of two scatterplots used in our study. The top row shows a strong average agreement score of $99.79\%$, while the bottom row has a low average agreement score below $50\%$.}
    \label{fig:vis_agreement}
\end{figure}

To illustrate our agreement score, \Cref{fig:vis_agreement} shows two stimuli with corresponding annotations and different agreement scores, whereby both stimuli having been annotated by five participants. In the top row, the five annotations demonstrate strong agreement in clustering, resulting in an average agreement score of $99.79\%$. In contrast, the bottom row shows annotations with an agreement score below $50\%$, where participants not only disagree on which points belong together, but one participant (most right) even indicated the absence of any clusters. Additionally, \Cref{fig:dataset_statistic} reveals that the occurrence of data samples decreases with an increasing number of clusters. For cluster numbers greater than six, the amount of data samples falls below $5\%$, which is too small a proportion to make general assumptions. Therefore, we exclude such stimuli from our training set and include only those highlighted by orange bars in \Cref{fig:dataset_statistic}.

\section{Evaluation}\label{sec:experiments}
For the evaluation of \methodName, we report performance for the test split of \ourDataset with the following dataset sizes: train ($1,171$ stimuli), validation ($87$ stimuli), and test ($206$ stimuli).

Similar to existing clustering algorithms like DBSCAN~\cite{ester1996density}, Birch~\cite{zhang1997birch}, and Optics~\cite{ankerst1999optics}, \methodName can predict on bivariate point sets directly, rather than rendered scatterplot images. As described in \Cref{sec:dataset}, the scatterplots used for crowdsourcing human annotations originate from real bivariate data. For the evaluation of our experiments, we compute the three metrics proposed in \Cref{sec:metrics} and report results averaged per group of cluster numbers from \Cref{fig:dataset_statistic}. To compensate for the imbalanced numbers of data samples per group, we compute the weighted average. Note that $\noiseIOU$ is computed only for data samples that contain points labeled as noise. Additionally, we provide a comprehensive evaluation of this experiment in Section 7  and Section 8 of the supplemental material, where we elaborate on further metrics like the Silhouette Index~\cite{rousseeuw1987silhouettes} and investigate the effect of dimensionality reduction on model performance.

\subsection{Outlier-Aware Rater Agreement}\label{sec:metrics}
Many measures exist that can quantify performance regarding different aspects. Therefore, selecting a metric that that can capture the intended solution to our problem is crucial. To evaluate \methodName, we need to measure the agreement between an isolated rater and a group of raters, which is computed by the Vanbelle Kappa Index~\cite{vanbelle2009agreement}, defined as:

\begin{equation}
    \vanbelle = \frac{\Pi_T - \Pi_E}{\Pi_M - \Pi_E}
\end{equation}
\noindent with $\Pi_T$ being the theoretical agreement, $\Pi_M$ the maximum attainable agreement and $\Pi_E$ the agreement expected by chance. Thus, the Vanbelle Kappa Index computes values in the range of $-1$ (minimal agreement) to $1$ (perfect agreement). A value of $0$ corresponds to the agreement expected by chance. Note that Vanbelle and Albert indicate in their work that if there is no variability in the rating from the isolated rater or from the group of raters, their index reduces to $\vanbelle=1$ for perfect agreement, and $\vanbelle=0$ otherwise when $\Pi_M = \Pi_E$. In cases where raters agree on \textsc{no cluster} and \textsc{single cluster}, all ratings are identical, leading to $\vanbelle=0$, if the group of raters does not perfectly agree, resulting in unreasonable scores computed by $\vanbelle$. In Section 13 of the supplementary material, we provide toy examples demonstrating such shortcomings of the Vanbelle Kappa Index.

Therefore, we develop a new measure that additionally accounts for outliers within the group of raters, and that can even measure if an isolated rating can improve agreement within the group. To do so, we define $\agree$ by averaging \Cref{eq:average_agreement_score} over all points and compute an averaged m-fold score $\cross$ for a given prediction $R$ and the group of raters $G$. We pick an annotation within the group and replace it with $R$, yielding a modified group $G^{R}$. Then we compute $\agree(G^{R})$ and repeat this $M$ times, where $M$ is equal to the number of annotations in the group. Thus, we compute an agreement index $\cross$, which indicates whether a given prediction improves agreement for $G$ when $\cross > 0$ or decreases it when $\cross < 0$. We define $\cross$ as follows:

\begin{equation}
    \cross(R, G) = \frac{1}{M}\sum^M{\agree(G^{R}_m) - \agree(G)}
    \label{eq:agreement_index}
\end{equation}

\noindent where $G$ is the group of annotations, and $G^{R}_m$ is the modified version of $G$ with the $m^{th}$ annotation replaced by $R$.

$\vanbelle$ and $\cross$ implicitly capture outliers; however, to explicitly measure noise detection, we further adopt a noise index $\noiseIOU$, which computes the average of an m-fold binary Jaccard Index $\mathfrak{J}$ over the group of raters and the prediction. This measure is defined as:

\begin{equation}
    \noiseIOU(\hat{R}, \hat{G}) = \frac{1}{M}\sum^M{\mathfrak{J}(\hat{R}, \hat{R}_{m})} 
    \label{eq:noise_iou}
\end{equation}

\noindent where $\hat{R}$ is the binary prediction for outliers, encoding points associated with noise as $0$ and points affiliating with a cluster as $1$. $\hat{R}_{m}$ is the binary annotation of the $m^{th}$ rater of group $G$.

Having defined these metrics, we will use them in the following sections to evaluate the results of \methodName.

\subsection{Contrastive Loss Weighting Analysis}\label{ssec:exp_weighting_factor}
In \Cref{subsec:loss}, we proposed our contrastive loss function $\mathfrak{L}_\text{mcl}$, along with weight matrix $\omega_{D}$, which is used to rescale the momentum of cases when two points are in dissimilar clusters. Depending on the scaling factor $D$, the weight matrix emphasizes gradients for dissimilar points, separating them into different clusters. In this experiment, we demonstrate how this can be used to adjust the behavior of our model using different values for $D \in [0.1, 1.0, 10.0, 50.0, 100.0]$. To do so, we train five models using the protocol from \Cref{ssec:train_protocol}, and we report performance in \Cref{tab:weight_factor_experiment}. Each model was trained for $37K$ steps, and looking at the results, the model trained with a weighting factor $D=50.0$ shows slightly better results for $\cross$ and $\vanbelle$, while noise prediction accuracy is behind the others. From the results presented in Section 2 of the supplementary material, we observe that the model separates clusters more distinctly for higher values of $D$ and merges clusters for smaller values. With medium values of $D=10$ and $D=50$, the clustering results are more variable, increasing clustering performance but also making it harder for the model to differentiate between noise and cluster decisions. However, in subsequent experiments, we show that this effect diminishes with a larger number of training steps, and we therefore use  $D=50.0$ in the remaining experiments unless stated otherwise. 

\subsection{Human Agreement Maximization}\label{ssec:exp_human_agreement_maximization}
In \Cref{ssec:human_cluster_results}, we discussed the agreement rate between raters, and \Cref{fig:dataset_statistic} shows that it is similar for different numbers of clusters. In this experiment, we investigate the impact of increasing human agreement during training on our model's performance. To do so, we use a threshold $T_\text{agree}$ to filter our training dataset. For each threshold, we discard training samples where the agreement score $\agree(G)<T_\text{agree}$. In this way, we construct five subsets using threshold values: $50\%,60\%,70\%,80\%,90\%$. In Section 12 of the supplemental material, we provide details on the selection of these threshold values. Then, for each subset, we train \methodName using the protocol from \Cref{ssec:train_protocol} and compare their performance results using \ourDataset. In this experiment, we use a fixed negative momentum $D=50.0$. 

The obtained results are shown in \Cref{tab:filter_agreement_experiment}. The first row shows results from the model trained on all available training data, \textsc{unfiltered}, and is identical to the best model from the first experiment (fourth row of \Cref{tab:weight_factor_experiment}). The results suggest, that while increasing the level of agreement in the annotations, the performance increases until reaching a maximum, as seen in row number four. From this point on, filtering for higher agreement rates reduces the number of training samples and decreases model performance. We conclude from this experiment, that a certain degree of variance in the annotations, helps to improve model performance, and that our model achieves robustness through such variability. As a result, we use the best model, trained with a threshold of $70\%$, for our remaining experiments, as shown in \Cref{tab:filter_agreement_experiment}.

\begin{table*}
\setlength{\tabcolsep}{2.5pt}
\centering
\footnotesize
\caption{Results of the contrastive loss weighting analysis (\emph{a}) in \Cref{ssec:exp_weighting_factor}, the human agreement analysis (\emph{b}) in \Cref{ssec:exp_human_agreement_maximization}, and the fine-tuning analysis (\emph{c}) in \Cref{ssec:exp_clustering_comparison}. We report performance results computing three metrics: $\cross, \vanbelle, \noiseIOU$ based on our test set.}

\begin{subtable}[t]{0.28\textwidth}
    \centering
    \begin{tabular}[t]{lccc}
    \toprule
    $D$ \hspace{6mm} &  \hspace{3mm}  $\cross$ \hspace{3mm}              & \hspace{3mm}   $\vanbelle$ \hspace{3mm}           & \hspace{3mm}  $\noiseIOU$ \hspace{3mm}          \\
    \midrule
      0.1      &   -5.64              &        0.53           &       49.97\%          \\
      1.0      &   -6.33              &        0.43           &       48.09\%          \\
     10.0      &   -2.01              &        0.61           &       46.26\%          \\
     50.0      &   \textbf{-1.66}     &        \textbf{0.62}  &       45.91\%          \\
    100.0      &   -1.78             &        0.61           &       \textbf{50.76\%}  \\
    \bottomrule
    \end{tabular}%
    \vspace{3.45mm}
    \caption{Analysis of different weights $D$ for negative samples in our contrastive loss}
    \label{tab:weight_factor_experiment}
\end{subtable}%
\hfill
\begin{subtable}[t]{0.38\textwidth}
    \centering
    \begin{tabular}[t]{lcccc}
    \toprule
    $T_\text{agree}$ & \hspace{1mm} \#samples \hspace{1mm}   & \hspace{2mm} $\cross$\hspace{2mm}      & \hspace{2mm} $\vanbelle$ \hspace{2mm}         &\hspace{2mm} $\noiseIOU$ \hspace{2mm}        \\
    \midrule
    \textsc{unfiltered}&   1171    &   -1.66          &        0.62           &       45.91\%          \\
     50\%              &   1148    &   -1.52          &        0.62           &       47.51\%          \\
     60\%              &   1049    &   -1.34          &        0.63           &       50.24\%          \\
     70\%              &    883    &   \textbf{-0.55} &        \textbf{0.68}  &       \textbf{58.48\%} \\
     80\%              &    672    &   -1.27          &        0.57           &       54.08\%          \\
     90\%              &    448    &   -1.66          &        0.58           &       53.92\%          \\
    \bottomrule
    \end{tabular}%
    \caption{Human agreement analysis by selecting different subsets of our data set} 
    \label{tab:filter_agreement_experiment}
    \end{subtable}%
    \hfill
\begin{subtable}[t]{0.3\textwidth}
    \centering
    \begin{tabular}[t]{lcccc}
    \toprule
    $D$ \hspace{6mm}    &    $T_\text{agree}$           & \hspace{2mm}  $\cross$ \hspace{2mm}                & \hspace{2mm}  $\vanbelle$ \hspace{2mm}          & \hspace{2mm} $\noiseIOU$ \hspace{2mm}  \\
    \midrule
     0.01          &       $70\%$             &   -0.85\%                &        0.66           &       57.07\%          \\
     0.1           &       $70\%$             &   \textbf{-0.52\%}       &        \textbf{0.69}   &      \textbf{60.91\%} \\
     1.0           &       $70\%$             &   -0.68\%                &        0.67           &       52.36\%          \\
    \bottomrule
    \end{tabular}%
    \vspace{10.42mm}
    \caption{Fine-tuning analysis of our best model by decreasing momentum for negative samples}
    \label{tab:finetune_experiment}
\end{subtable}
\vspace{2mm}
\end{table*}

\subsection{Clustering Techniques Comparison}\label{ssec:exp_clustering_comparison}
In this experiment, we investigate the gap between existing clustering techniques and \methodName. Therefore, ten state-of-the-art clustering algorithms are compared using the implementation from scikit-learn~\cite{pedregosa2011scikit}. Since all approaches require parameterization, we first conduct a parameter search using our training dataset to find the optimized parameters for each technique. Second, we fine-tune \methodName using the insights from \Cref{ssec:exp_weighting_factor} and \Cref{ssec:exp_human_agreement_maximization}, investigating whether smaller values of $D$ improves single cluster predictions. We train three different models using momentum values $D=[0.01, 0.1, 1]$. In \Cref{tab:finetune_experiment}, we report the performance results for all three models. Since the model has already learned to separate clusters, fine-tuning it to reduce its cluster separation ambitions must be done carefully. Therefore, we fine-tune each mode using a reduced learning rate of $10^{-7}$ for $8K$ training steps. The results show that fine-tuning our model using a weighting factor $D=0.1$ helps to improve single cluster predictions and even the overall performance of our model. In Section 8 of the supplemental material, we evaluate the performance of such a model for specific numbers of clusters, separately. We demonstrate that the largest error contribution originates from predictions for single clusters. As a baseline for our evaluation, we adopt an image-based model utilizing a pre-trained image segmentation U-Net~\cite{ronneberger2015u}, which we fine-tune it on our collected scatter plot stimuli. Details on the training protocol can be found in Section 4 of the supplementary material.

We then evaluate each approach using \ourDataset and compare it against \methodName using the best model from the fine-tuning experiment. We report performance results in the upper part of \Cref{tab:clustering_technique_experiment}. Note that some techniques do not compute outliers, for which we do not calculate $\noiseIOU$. The baseline model performs better for the $\cross$ metric compared to the standard clustering techniques that do not use prior knowledge about the number of clusters. This indicates that predictions are well-aligned with human ratings. The results for $\vanbelle$ and $\noiseIOU$ indicate average alignment compared to existing techniques. Looking at the results for \methodName, we can see that it outperforms all compared state-of-the-art approaches on all metrics. The $\cross$ measure shows that predictions from \methodName almost perfectly agree with the group of raters, indicated by a value close to zero. The $\vanbelle$ also suggests superior performance of \methodName over the other clustering techniques, which we interpret as indicating that \methodName predictions agree with human annotators. Examining the closest competitors: Ward, K-means, and Gaussian Mixture Models, which have similar $\vanbelle$ scores compared to \methodName, we see that traditional clustering algorithms can be used to reflect human judgments. This is also supported by $\cross$ values close to zeros; however, these methods require a prior of known number of clusters, which is inferred automatically by our method. Finally, examining the accuracy for noise or outlier prediction, \methodName has a slight advantage over DBSCAN, while both approaches outperform Optics. Additionally, we measured the running time for inference of cluster predictions utilizing the test split of \ourDataset for 206 inference steps resulting in 5.16 sec. That translates to 0.025 sec per step utilizing an Nvidia RTX 3060 and batch size of 64. As comparison DBSCAN has a running time of 3.48 sec or 0.017 sec per step.

\begin{table*}[ht!]
\setlength{\tabcolsep}{2.5pt}
\centering
\footnotesize
\caption{Comparison between \methodName, the image baseline and ten state-of-the-art clustering techniques using two test datasets: \ourDataset(\Cref{ssec:exp_clustering_comparison}) and \sedlmair~(\Cref{ssec:exp_sdr}). We do not compute $\noiseIOU$ for cluster techniques which do not compute outliers. Results highlighted with a * are from clustering techniques that require priors about the number of clusters as per the ground truth annotation.}
\begin{tabular}{ccccccccccccc}
\toprule
&\methodName & Baseline    &  DBSCAN  & OPTICS &  Ward & Mean       & Affinity    & Spectral   &  Agglomerative & BIRCH & K-means &  Gaussian Mixture    \\
&            &             &          &        &       & Shift      & Propagation & Clustering &  Clustering    &       &         &  Model               \\
&            &             &  \cite{ester1996density}        &    \cite{ankerst1999optics}    &    \cite{murtagh2011ward}   & \cite{comaniciu2002mean}      & \cite{frey2007clustering} & \cite{ng2001spectral} &   \cite{moseley2017approximation}   &   \cite{zhang1997birch}    &     \cite{hartigan1979algorithm}    &         \cite{bensmail1997inference}         \\
\midrule
\multicolumn{13}{c}{Evaluation on our test split}\\
\midrule
 $\cross	\uparrow$       &    \textbf{-0.52}       & -2.40   &  -4.13          &          -5.56          &    -1.04\textsuperscript{*}               &  -6.90     &   -10.34    &    -2.65\textsuperscript{*}        &       -7.02    &            -7.68       &         -1.49\textsuperscript{*}             &          -0.90\textsuperscript{*}        \\
 $\vanbelle	\uparrow$   &    \textbf{0.69}        & 0.51    &   0.62          &           0.56          &     0.66\textsuperscript{*}               &   0.53     &     0.39    &     0.53\textsuperscript{*}        &        0.38    &             0.38       &          0.63\textsuperscript{*}             &           0.67\textsuperscript{*}        \\
 $\noiseIOU	\uparrow$   &    \textbf{59.67\%}     & 36.13\% &   52.12\%       &           21.29\%       &    -                                      &     -      &        -    &     -                              &       -        &             -          &          -                                   &             -                             \\
 \midrule
 \multicolumn{13}{c}{Evaluation on \sedlmair data set.}\\
\midrule
 $\cross	\uparrow$       &   \textbf{-1.34}        & -5.35  &  -3.19          &          -3.15          &    -7.21\textsuperscript{*}                &  -5.28   &   -13.04    &    -7.21\textsuperscript{*}  &       -5.85    &    -6.47       &    -7.21\textsuperscript{*}   &   -7.21\textsuperscript{*}        \\
 $\vanbelle	\uparrow$   &    \textbf{0.64}        & 0.43   &   0.50          &           0.60          &     0.62\textsuperscript{*}                &   0.60   &     0.31    &     0.62\textsuperscript{*}  &        0.43    &     0.46       &      0.62\textsuperscript{*}  &     0.62\textsuperscript{*}        \\
 $\noiseIOU	\uparrow$   &   \textbf{46.28\%}      & 33.86\% &   28.36\%       &          36.11\%        &    -                                       &     -    &        -    &     -                        &       -        &       -        &          -                    &             -                      \\
 \bottomrule
\multicolumn{13}{l}{*The ground truth number of cluster was given to compute these scores.}\\
\label{tab:clustering_technique_experiment}
\end{tabular}%
\end{table*}

\begin{figure}[t]
    \centering
    \includegraphics[width=\linewidth]{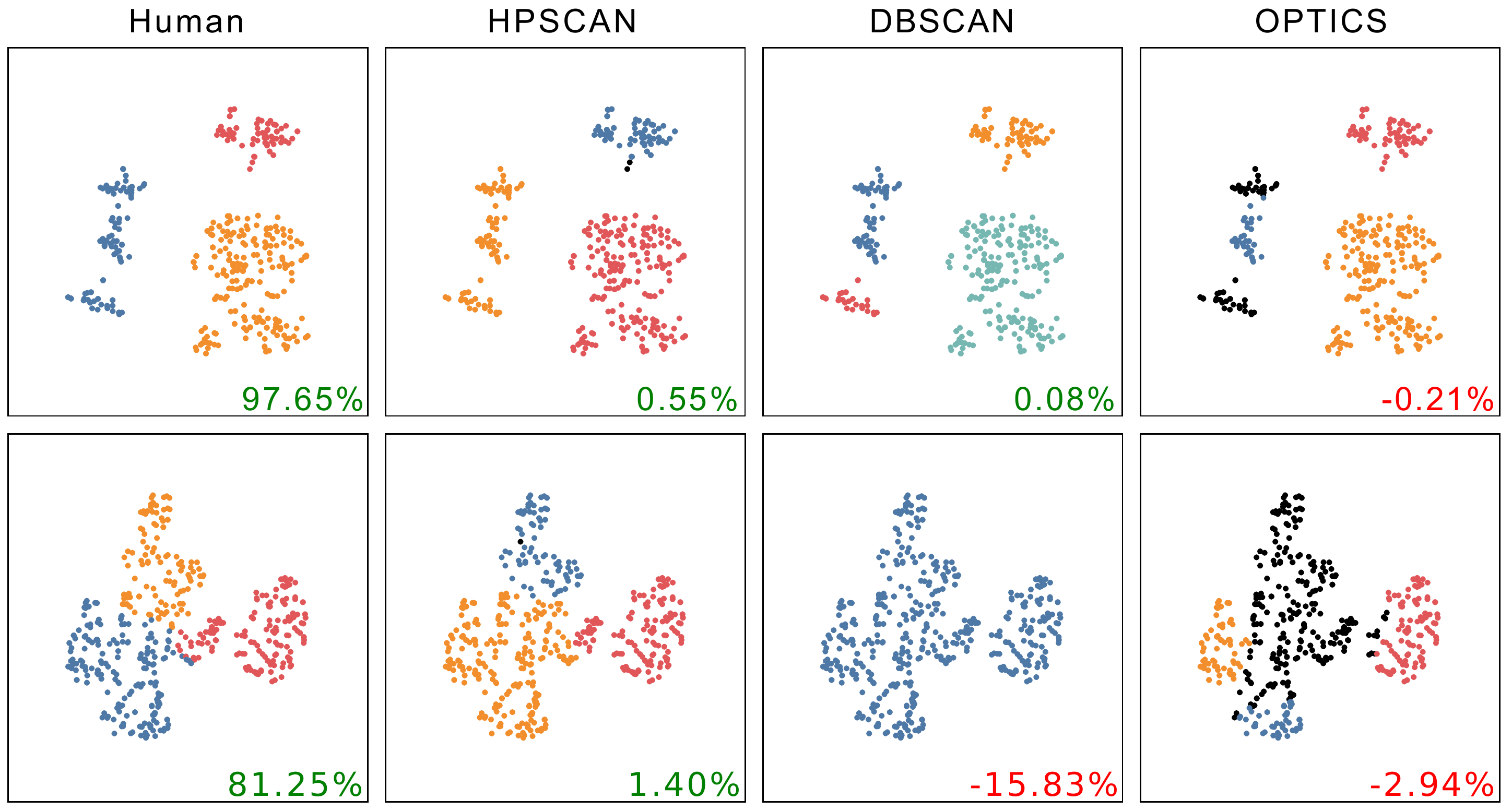}
    \caption{Qualitative evaluation results: Two point sets (rows) sampled from the \sedlmair dataset are shown. Each column presents outcomes for different clustering techniques. We compare \methodName to the two best performing standard techniques, along with human ratings that display the highest agreement among raters, shown in the first column. For each technique, the corresponding $\cross$ index is shown in \textcolor{ForestGreen}{green}, when the clustering improves group agreement, and in \textcolor{red}{red} otherwise.}
    \label{fig:qualitative_results}
\end{figure}

\subsection{Generalization Analysis}\label{ssec:exp_sdr}
In this experiment, we aim to investigate the ability of \methodName to generalize to unseen data. To this end, we utilize the dataset provided by the work of Sedlmair~\cite{sedlmair2013empirical}, which consists of 2D scatterplots derived using dimensionality reduction techniques (SDR) and includes both real and synthetic data. We apply the same crowdsource annotation procedure as described in \Cref{sec:dataset} to collect clustering annotations from $68$ participants for $272$ stimuli. We then evaluate the best model from \Cref{ssec:exp_clustering_comparison} using this dataset and report results for inferring clustering in \Cref{tab:clustering_technique_experiment} and for regressing agreement predictions in \Cref{tab:uncertainty_regression_results}. We demonstrate that \methodName outperforms existing clustering techniques on all three measures, $\cross, \vanbelle, \noiseIOU$ indicating better alignment with human ratings and showcasing the ability of \methodName to generalize to unseen data. Qualitative results for \methodName generated during inference are shown in the upper row of \Cref{fig:qualitative_results}. These estimated clusterings for the \sedlmair dataset underline the well-aligned predictions of \methodName with human judgments, and we refer the reader to Section 5 of the supplemental material for further evaluation results.


\subsection{Human Agreement Estimation}\label{ssec:exp_uncertainty}
\begin{figure}
    \centering
    \includegraphics[width=\linewidth]{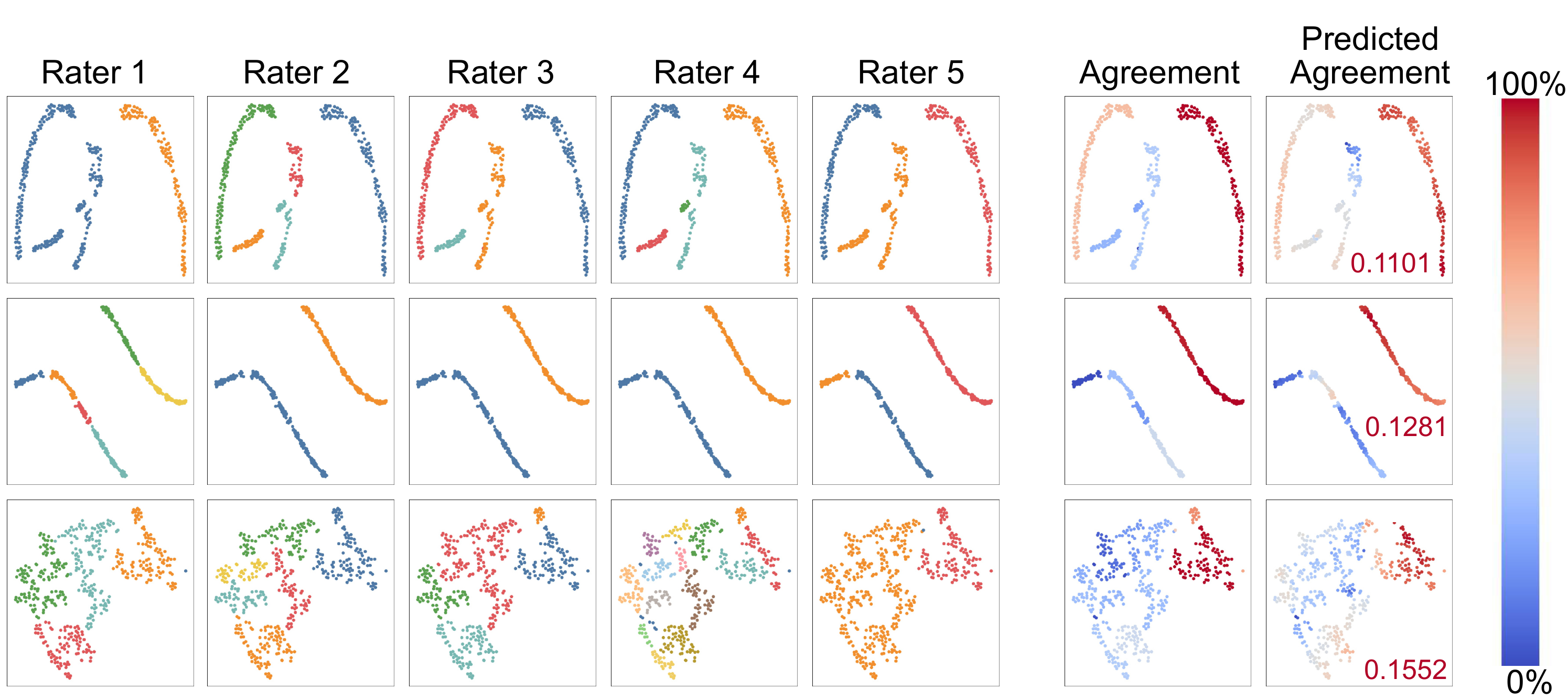}
    \caption{The first five columns display human annotations collected during our online crowdsource study. We compare annotations for three stimuli. In the sixth column, the computed agreement score per point is shown. Finally, in the last column, the prediction of \methodName is shown, along with the averaged absolute error over all points.}
    \label{fig:uncertainty}
\end{figure}

After investigating the effect of maximizing human agreement during training to learn a robust model in \Cref{ssec:exp_human_agreement_maximization}, one question remains: what if human raters do not agree on a clustering? On the left side of \Cref{fig:uncertainty}, we display clusterings proposed by five human annotators for the same stimulus, as well as the corresponding agreement rate amongst the group of raters, shown in the sixth column. 
Consequently, we extend \methodName to also output an agreement score $\agree$, as proposed in \Cref{eq:average_agreement_score}, for each point. The estimation of our network has the shape $N \times 1$ and is computed by applying a Sigmoid activation to the network output. We use the same fine-tuning procedure as described in \Cref{ssec:exp_clustering_comparison}, with identical hyperparameters, and train \methodName on \ourDataset. We then evaluate it on the test split of \ourDataset and \sedlmair, reporting regression results measured in Mean Squared Error (MSE) and Mean Absolute Error (MAE) in \Cref{tab:uncertainty_regression_results}. Looking at the results, there appears to be no large gap between the two datasets considering the regression values, and together with qualitative evaluation results, there is an indication that \methodName is capable to reflect human agreement with an error of 13 points in MAE for \ourDataset and 15 points in MAE for \sedlmair dataset.
In the last column of \Cref{fig:uncertainty}, we display prediction results of \methodName for the \sedlmair dataset.
\begin{table}[h]
    \centering
    \setlength{\tabcolsep}{2.5pt}
    \begin{tabular}{lcc}
    \toprule
    Test dataset     \hspace{2mm}                &\hspace{2mm} MSE \hspace{2mm}   & \hspace{2mm} MAE \\
    \midrule
    \methodName                       & 0.0305 & 0.1393 \\
    \sedlmair                         & 0.0317 & 0.1564 \\
    \bottomrule
    \end{tabular}
    \caption{We evaluate our model, that is trained on our dataset, for the human agreement estimation task. We compute two regression metrics to measure its performance on two datasets: \methodName and the dataset \sedlmair, that consists of real and synthetic scatterplots.}
    \label{tab:uncertainty_regression_results}
\end{table}

\subsection{Cluster Counting Experiment}\label{ssec:exp_clustme}
In this experiment, we apply \methodName to the ClustMe dataset~\cite{abbas2019clustme}, which consists of $1,000$ stimuli with human judgments available. Each stimulus was annotated by $34$ human raters to determine whether one or more than one cluster is perceived by the participant. We collect predictions from \methodName for each stimulus and convert the clustering result into the estimated number of clusters. We compute the performance results of \methodName using the Vanbelle Kappa Index, accuracy, and F1 score, and report the results in \Cref{tab:cluster-counting}. To compute accuracy and F1 score, the majority cluster count from the ground truth was used as a binary label. The results indicate poor performance results compared to the approaches presented in ClustMe. To contextualize the results of \methodName, we show the performance of three dummy models: the random cluster model, which estimates randomly between single and multiple clusters; the single cluster model, which always estimates a single cluster; and the multi-cluster model, which always estimates more than one cluster. Indicated by the accuracy of $80\%$ for the single cluster model and $20\%$ for the multi-cluster model, ClustMe has a class imbalance favoring single clusters. Additionally, ClustMe datasets are generated using arbitrarily configured Gaussian Mixture Models (GMM). Consequently, \methodName achieves a Vanbelle Kappa Index of $0.5$, an accuracy of $84\%$, and an F1 score of $55\%$, suggesting a tendency to estimate mainly single clusters for ClustMe. This experiment reveals a domain gap between dataset distributions originating from projections through dimensionality reduction and GMM-based scattered data.

\begin{table}[]
    \centering
    \begin{tabular}{lccc}
    \toprule
    model & $\vanbelle$ & Acc & F1 \\
    \midrule
    \methodName & $\textbf{0.50}$ & $\textbf{84\%}$ & $\textbf{55\%}$ \\ 
    $\Theta_{random}$ & $0.00\pm0.03$ & $50\pm2\%$ & $28\pm2\%$ \\
    $\Theta_{single}$ & $0.00$       & $80\%$ & $0\%$ \\
    $\Theta_{multi}$  & $0.00$       & $20\%$ & $33\%$ \\
    \bottomrule
    \end{tabular}
    \caption{Evaluation results on the ClustMe\xspace\cite{abbas2019clustme} dataset, which consists of $1,000$ stimuli annotated by $34$ human raters. Each annotation is a binary decision whether a single cluster or multiple cluster were perceived by the human. }
    \label{tab:cluster-counting}
\end{table}

\section{Discussion and Future Work}\label{sec:limitations}
We demonstrate \methodName, a clustering model based on human-perceived clustering annotations. We collected a large-scale dataset using Prolific to crowdsource human point-wise clustering annotations. Investigation of the collected data shows agreement above $80\%$ between human subjects. This dataset enables us to train \methodName, a learned clustering model that mimics point clustering as performed by the human visual system. In multiple experiments, our protocol for training a point-based model is demonstrated, and we show how we fine-tune our model to adjust it to human annotations. To evaluate \methodName, we further proposed a novel metric, that measures agreement improvement while also being sensitive to annotation consistency. Ultimately, we evaluate our model using \sedlmair and \datagov, which are datasets featuring unseen and out-of-scope data, and compare it against ten state-of-the-art clustering techniques, with \methodName outperforming all compared algorithms, demonstrating its ability to generalize to new data. In our experiments, we also present how to deal with multiple correct clusterings, e.g., when multiple human judgments disagree on a particular stimulus. \methodName can be used to estimate human agreement, enabling the detection of such ambiguities.

While \methodName is the first of its kind and performs well on human-like clustering, it has some limitations. PointNet++ is order invariant with respect to the input point cloud, and our contrastive loss term is invariant to the order of cluster IDs. However, constructing the similarity matrix has a complexity of $O(n^2)$. In our experiments, we used a fixed size for point clouds, $N=512$, to keep computational costs low. As a result, we had to discard or randomly sample from real-world datasets, to match such requirements. In the field of contrastive learning, approaches like SimCLR~\cite{chen2020simple} or SwAV~\cite{caron2020unsupervised} exist, which do not have this limitation and could be used to cluster points based on their latent codes. Note that during training time, \methodName is limited by a fixed size of $N$, but during inference, PointNet++ and thus our approach as well can process arbitrary numbers of points, as demonstrated in the evaluation of \sedlmair dataset in \Cref{ssec:exp_sdr}. During training, we set $C$ to 20, which is also the maximum number of clusters that HPSCAN can detect. As a result, our model is currently unable to handle datasets with more than 20 clusters. However, based on our observations, it is unlikely that a human judge would perceive more than this number of distinct clusters in a typical scatterplot, so we consider this limitation reasonable for the intended applications.

\methodName is trained on human-annotated data collected during a crowdsourced study. Therefore, we rendered datasets using a fixed visual encoding (marker size, opacity, color, etc.), which limits the conclusions about the agreement and perceived numbers of clusters to such visual encoding. In cases where visual encoding affects cluster perception, this would not be captured by \methodName. However, as shown by Quadri\etal\cite{quadri2022automatic} visual encodings interfere with perceived clustering, and clusterings proposed by HPSCAN might look inaccurate if, for instance, larger markers are used in the scatterplot. Nevertheless, if desired, our crowdsourced study could have also included variable visual encodings, and provided it as point features to our network during training. Thus, this is not strictly a limitation of \methodName, as more stimuli data could incorporate visual encodings during cluster prediction.

\methodName is developed as an alternative to existing clustering techniques, specifically aligned to human perception. However, this work fixes the visual encoding during dataset construction in order to keep the design space manageable. Therefore, exploring the effect of visual encoding is subject to future research, while this work provides a promising baseline that is able to generalize beyond unseen data. Further, our analysis of human agreement on clustering shows a large variety of agreement degree depending on the structure of the underlying data. It is questionable how meaningful a single clustering prediction is for cases when a group of human raters provides ambiguous annotations. A generative model could be trained to produce many different such human-aligned clusterings, reflecting the same human ambiguous distribution of annotations. In this work, we decided to predict an agreement score for each point, providing neural feedback about the learned human discord. For a user of \methodName, this provides some understanding of the reliability of the corresponding estimated clustering.

While we see \methodName as a milestone for perception-based clustering, we see several endeavors for future research. First, we could investigate the influence of visual stimuli on cluster perception. Furthermore, we could leverage the presented technology to learn other scatterplot tasks, such as noise detection. Third, there is room for future incorporation of GMM-based data during training to achieve more robustness on synthetic data. Finally, we would like to investigate how \methodName can be used to optimize scatterplot visualization parameters along the lines of Wang~\etal\cite{wang2017perception,wang2018optimizing}.

\section{Data Availability Statement}
The data that support the findings of this study are openly available in github.com at \url{https://github.com/kopetri/HPSCAN}.

\printbibliography                

\includepdf[pages=-]{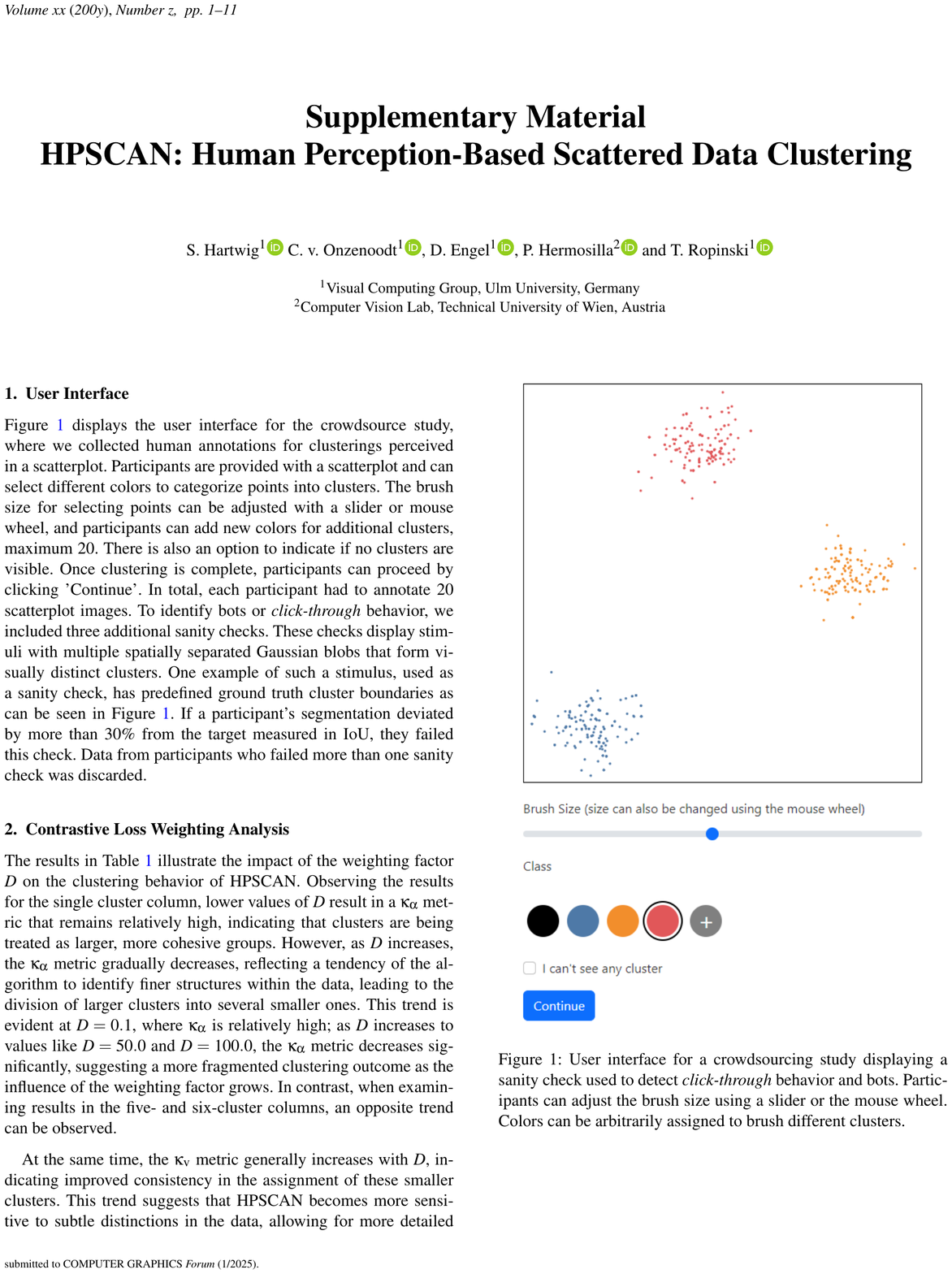}
\end{document}